\documentclass{bvm} % Do NOT change this line

\usepackage{tikz}
\newcommand\PublicationNote[2]{%
\begin{tikzpicture}[remember picture,overlay]
\node[anchor=north,draw,yshift=-#1] at (current page.north) {\parbox[t][][c]{\linewidth}{\small#2}};
\end{tikzpicture}%
}
\begin{document}
% Do not change anything in the preamble (anything above \begin{document}) except for the specification of the bibliography file, any additional changes will be lost

% You can create your own commands using \newcommand. These must be placed after(!) \begin{document} and should contain your own name to avoid multiple definitions between different papers
\newcommand{\bvmyear}{2025}

% use the \selectlanguage command to select the language in which your proceedings are written
%\selectlanguage{ngerman} % German
\selectlanguage{english} % English

% The title of your paper
%\title{Weakly Supervised Segmentation of HRF in Retinal OCT}
\title{Weakly Supervised Segmentation of HRF in OCT with Compact Convolutional Transformers and SAM~2}
% If you write a short paper/abstract, the title must start with "Abstract:".
% \title{Abstract: Bildverarbeitung für die Medizin \bvmyear}

% Optional specification of a subtitles
\subtitle{}

% titlerunning appears in the header of every second page
% LaTeX generates this automatically from your paper title
% However, if it is too long, the message "Title Suppressed Due to Excessive Length" appears instead.
% In this case, specify an abbreviated form of the title here
\titlerunning{Weakly Supervised Hyper-Reflective Foci Segmentation}

% Please indicate all authors involved
% To allow us to correctly identify the last name of each author, indicate it using the \lname{} command.
% If more than one institute is involved, list the number of the institute(s) (see below) with \inst{} after the respective author. If only one institute is involved, omit this.
% Separate all authors with a comma
\author{Olivier \lname{Morelle} \inst{1,2}, 
	Justus \lname{Bisten} \inst{1}, 
        Maximilian~WM. \lname{Wintergerst} \inst{2,5},
        Robert~P. \lname{Finger} \inst{2,4},
	Thomas \lname{Schultz} \inst{1,3},
}

% Enter the authors here as you want them to appear in the header
% Name only the surnames
% Depending on the number of authors involved, follow the examples below
% \authorrunning{Meier} - one author
% \authorrunning{Meier \& Müller} - two authors
% \authorrunning{Meier, Müller \& Schulze} - three authors
% \authorrunning{Meier et al.} - more than three authors
\authorrunning{Morelle et al.}

% Specify the institutes involved
% In case of participation of more than one institute, each institute shall be preceded by an ascending number with \inst{}.
% If only one institute is involved, omit the corresponding number.
% Separate individual institutes with \\
\institute{
\inst{1} B-IT and Department of Computer Science, University of Bonn \\
\inst{2} Department of Ophthalmology, University Hospital Bonn \\ 
\inst{3} Lamarr Institute for Machine Learning and Artificial Intelligence \\
\inst{4} Department of Ophthalmology, University Medical Center Mannheim, Heidelberg University \\
\inst{5} Augenzentrum Grischun, Chur, Switzerland}

% Enter the e-mail address of the corresponding author
\email{schultz@cs.uni-bonn.de}

\maketitle

% Abstract of your paper, only for long papers
% Do NOT use \begin{abstract} ... \end{abstract} for short articles
\begin{abstract}
Weakly supervised segmentation has the potential to greatly reduce the annotation effort for training segmentation models for small structures such as hyper-reflective foci (HRF) in optical coherence tomography (OCT). However, most weakly supervised methods either involve a strong downsampling of input images, or only achieve localization at a coarse resolution, both of which are unsatisfactory for small structures. We propose a novel framework that increases the spatial resolution of a traditional attention-based Multiple Instance Learning (MIL) approach by using Layer-wise Relevance Propagation (LRP) to prompt the Segment Anything Model (SAM~2), and increases recall with iterative inference. Moreover, we demonstrate that replacing MIL with a Compact Convolutional Transformer (CCT), which adds a positional encoding, and permits an
exchange of information between different regions of the OCT image, leads to a further and substantial increase in segmentation accuracy.

\end{abstract}

\PublicationNote{1cm}{This is a preprint of the following paper: Morelle et al., Weakly Supervised Segmentation of Hyper-reflective Foci with Compact Convolutional Transformers and SAM 2, published in Bildverarbeitung für die Medizin 2025, edited by Palm, C., et al., 2025, Springer Vieweg, reproduced with permission from Springer Fachmedien Wiesbaden GmbH. The final authenticated version is available online at: \href{http://dx.doi.org/10.1007/978-3-658-47422-5_23}{http://dx.doi.org/10.1007/978-3-658-47422-5\_23}.}

\section{Introduction}

Age-related macular degeneration (AMD) is a leading cause of vision loss with age, characterized by the deterioration of the central retina~\cite{liPrevalenceIncidenceAgerelated2020}. Optical coherence tomography (OCT) provides high-resolution cross-sectional images of the retina. Within these OCT images, hyper-reflective foci (HRF) appear as small, bright spots and are considered biomarkers for AMD progression~\cite{vermaRelationshipDistributionIntraretinal2023}. Segmentation of HRF is crucial for assessing disease severity and monitoring treatment efficacy.

However, manual segmentation of HRFs is a labor-intensive and time-consuming task, particularly due to the high resolution of the images and the minute size of HRFs. Our dataset comprises images of size $496 \times 1024$ pixels, with HRFs exhibiting a median size of only 17 pixels. This makes weak supervision an attractive alternative.

Weakly supervised segmentation, which achieves pixel-level localization based on image-level annotations, has gained increasing attention for OCT data. One method proposed generating segmentation maps from image-level labels using multi-scale class activation maps~\cite{maMSCAMMultiScaleClass2020a}. Another one focused on anomalous structures, which it segmented without requiring detailed annotations~\cite{wangWeaklySupervisedAnomaly2021}. The TSSK Net employed a teacher-student architecture to enhance segmentation performance under weak supervision~\cite{liuTSSKNetWeaklySupervised2023}. It has been employed for the segmentation of various biomarkers in OCT, including HRFs, but its architecture involves a downsampling of images to $256 \times 256$ pixels. At that resolution, 16\% of the annotated HRFs in our dataset are lost. Working with most Vision Transformer (ViT) based methods, such as RETFound \cite{zhouFoundationModelGeneralizable2023}, even requires downsampling to $224 \times 224$ pixels, which loses 22\% of our annotated HRFs. Since we consider such large losses of diagnostically relevant detail unacceptable, we investigate methods for weak supervision that preserve the full resolution.

The first contribution of our work is a framework for weakly supervised HRF segmentation based on Multiple Instance Learning (MIL), which classifies full-resolution images by treating them as a bag of patches. Attention-based MIL \cite{ilseAttentionbasedDeepMultiple2018} improves accuracy, and enables a certain level of localization, by computing an attention score per patch. Since HRFs are often much smaller than patches, we combine that approach with Layer-wise Relevance Propagation (LRP)~\cite{bachPixelWiseExplanationsNonLinear2015}, leading to relevance maps that highlight the contributions of individual pixels to the model's predictions.

We achieve a final segmentation by using the relevance map to prompt version~2 of the Segment Anything Model (SAM~2)~\cite{raviSAM2Segment}, a powerful foundation model capable of producing accurate segmentation masks based on minimal input prompts. We derive a prompt from the most relevant pixel, developing a suitable prompting strategy that compensates for the fact that SAM~2 is not well-suited for small structures such as HRFs. Since a single image can contain more than one HRF, we iteratively pass it through the pipeline, occluding all previously detected HRFs, until no further HRFs are found. 

Our second contribution is to investigate the relative benefit of replacing MIL with a Compact Convolutional Transformer (CCT) \cite{hassaniEscapingBigData2022}, a transformer architecture that is well-suited for small datasets, and flexible with regard to input resolution. We hypothesized that the CCT model would outperform the MIL model due to its ability to capture positional information and facilitate information exchange between patches through attention mechanisms, reducing the likelihood of missing small HRFs on patch borders.

\section{Materials and methods}

\subsection{Data}

This study utilized 191 OCT volumes from the Laser Intervention in Early Stages of Age-Related Macular Degeneration (LEAD) study \cite{guymerSubthresholdNanosecondLaser2019}, all imaged using Heidelberg Spectralis devices and manually annoted for hyperreflective foci (HRF) by three raters \cite{gohHYPERREFLECTIVEFOCINOT2024}. Due to the poor between-rater reproduction of very small annotations (less than 5 pixels), we excluded them from the analysis. Each volume consists of multiple slice images (B-scans), leading to 962 B-scans with HRF and 8,392 B-scans without HRF. Overall 2,146 HRFs where annotated with an average of 12 per volume. To ensure robust evaluation, the dataset was divided into training and test sets using an 80/20 split, stratified by HRF count per volume to maintain similar distributions and prevent information leakage. An additional 20\% of the training set was reserved for validation during model training. 

\subsection{Image-level predictions}

We investigated the relative benefits of different forms of weak supervision by reducing the pixel-level annotations to three types of image-level labels which contain increasing amounts of information, at an increasing expected annotation effort: In a \textbf{binary classification} task, we only differentiated between B-scans with and without HRF. In \textbf{multi-class classification,} we categorized B-scans into no HRF, one HRF, or more than one HRF. In a \textbf{regression approach,} we trained a model to estimate the number of HRFs in a given B-scan, with labels clipped to a maximum of 10 HRFs per B-scan to focus on values with sufficient support in our data.

\paragraph{Model Architectures}

Two architectures that can process the B-scans at full resolution were employed for image-level predictions: a Multiple Instance Learning (MIL) model and a Compact Convolutional Transformer (CCT). The regression variant of both models used a sigmoid-activated output, scaled by a factor of 10.

The \textbf{MIL model} consisted of a bag of AlexNets with an input shape of \(64 \times 64\) pixels. Inspired by Ilse et al.~\cite{ilseAttentionbasedDeepMultiple2018}, instances were weighted and averaged via self-attention. The model was initialized with ImageNet weights from \texttt{torchvision} and trained using a batch size of 64 for 50,000 steps, following a OneCycle learning rate policy with a maximum learning rate of \(1 \times 10^{-6}\). The optimizer used was AdamW with a weight decay of 0.01. The retinal region in the input images was localized using Otsu thresholding, and 3 rows of patches were extracted from this region.

The \textbf{CCT model}, based on Hassani et al.~\cite{hassaniEscapingBigData2022}, is designed for small datasets and integrates convolutional layers for tokenization. This approach captures local spatial relationships and preserves high-resolution details, beneficial for tasks requiring precise localization. The model was trained from scratch with a batch size of 16 for 50,000 steps, using a OneCycle learning rate policy with a maximum learning rate of \(1 \times 10^{-5}\) and the AdamW optimizer with a weight decay of 0.01.

\paragraph{Data Augmentation}

During training, data augmentation techniques were applied, including random vertical shifts, horizontal flips, scaling, rotation, and intensity adjustments. Images were normalized using the mean and standard deviation of the training dataset.

\subsection{Weakly supervised segmentation}

For weakly supervised segmentation, Layer-wise Relevance Propagation (LRP) \cite{bachPixelWiseExplanationsNonLinear2015} was implemented for both models. For the CCT, this involved a recent extension of LRP to transformer architectures \cite{Ali-LRP-transformer}. Based on the resulting relevance maps, a square, fixed-size bounding box around the most relevant pixel was used as a prompt for version~2 of the Segment Anything Model (SAM~2) \cite{raviSAM2Segment}. Since we found that SAM~2 is not well-suited for the segmentation of small objects such as HRFs at the native resolution, we crop the B-scan, centered on the position of the prompt, and subsequently upsample the resulting patch to SAM's input resolution, effectively making the HRF sufficiently large for reliable segmentation. The most suitable crop and bounding box sizes were established using a grid search over all HRFs from the training data. A box size of 4 pixels and crop sizes between 50 and 100 pixels yielded optimal results, achieving a mean Dice around 0.67 when prompting with ground truth HRF centers.

The highest scoring mask returned by SAM is selected. Optionally, an attempt is made to segment additional HRFs with an iterative procedure that inpaints the previously detected HRFs with the image mean and re-runs the model until the predicted value falls below a threshold of 0.05, or a maximum number of iterations is reached. A comparison of several inpainting methods indicated that simply replacing HRFs with the image mean was most effective.

\section{Results}
We evaluated the performance of the CCT and MIL models as binary classifiers and weakly supervised segmentation models. For both models 3 variants where trained to study the effect of additional information during the training on the performance.

\begin{table}[t]
\centering
\caption{Classification and segmentation metrics, with and without iterative inference, for CCT and MIL models. The 2-classes task is a binary classification for the presence of HRF. For the 3-classes task, the classes are no HRF, 1 HRF and more than 1 HRF. For the regression task, the target is the HRF count in the B-scan, clipped to a maximum of 10.}
\label{3696-tab:model_performance}
\begin{tabular*}{\textwidth}{l@{\extracolsep\fill}rccccccccc}
\hline
\textbf{Model}                & \multicolumn{1}{c}{\textbf{Task}} & \multicolumn{3}{c}{\textbf{Classification}}         & \multicolumn{3}{c}{\textbf{Segm. 1 Iteration}}   & \multicolumn{3}{l}{\textbf{Segm. 6 Iterations}}  \\ \cline{3-11} 
                              &                                   & \textbf{AUROC} & \textbf{Avg. Prec} & \textbf{F1}   & \textbf{Dice} & \textbf{Recall} & \textbf{Preci} & \textbf{Dice} & \textbf{Recall} & \textbf{Preci} \\ \hline
\multirow{3}{*}{\textbf{CCT}} & 2-Classes                         & 0.90           & 0.70               & 0.64          & \textbf{0.33} & \textbf{0.35}   & \textbf{0.47}  & \textbf{0.33} & \textbf{0.42}   & \textbf{0.37}  \\
                              & 3-Classes                         & \textbf{0.92}  & \textbf{0.74}      & \textbf{0.65} & 0.29          & 0.30            & 0.41           & 0.31          & 0.40            & 0.36           \\
                              & Regression                        & 0.90           & 0.58               & 0.6           & 0.22          & 0.25            & 0.29           & 0.21          & 0.35            & 0.22           \\
\multirow{3}{*}{\textbf{MIL}} & 2-Classes                         & 0.86           & 0.58               & 0.53          & 0.11          & 0.09            & 0.23           & 0.13          & 0.12            & 0.23           \\
                              & 3-Classes                         & 0.86           & 0.58               & 0.51          & 0.08          & 0.06            & 0.19           & 0.08          & 0.07            & 0.18           \\
                              & Regression                        & 0.85           & 0.42               & 0.41          & 0.05          & 0.04            & 0.13           & 0.07          & 0.07            & 0.13           \\ \hline
\end{tabular*}
\end{table}

\paragraph{Classification}

As presented in Tab. \ref{3696-tab:model_performance}, the CCT model consistently outperforms the MIL model across all metrics and for all training regimes. The best CCT model for the classification task was trained with 3-Classes and is only slightly better than the one trained with two classes. It achieves an AUROC of 0.92, an Average Precision of 0.74 and an F1 score of 0.65. For the MIL model there is also not much difference between training with 2 or 3 classes. But for both models the regression task performs worst.

\paragraph{Segmentation}

For the weakly supervised segmentation task, the CCT trained with 2 classes yields the best results. The difference in performance between the MIL model and the CCT is even more pronounced than for the segmentation task. Iterative inference increases the (pixel-level) recall, but also reduces precision, and therefore does not yield a substantial benefit in terms of Dice. This is mostly due to the fact that averages in Tab. \ref{3696-tab:model_performance} include a large number of B-scans with only one HRF. Images that contain more than one HRF are more challenging to segment (average Dice with the 2-class CCT after the first iteration is only 0.22), but benefit more clearly from the iteration (Dice increases to 0.28 after three iterations).

\paragraph{Relevance processing}

We evaluated multiple approaches to obtain segmentation masks from the classifiers relevance maps shown in Tab. \ref{3696-tab:SAM_Ablation}. A global threshold calibrated on the validation data is straight forward but results in a Dice of only 0.16 for the best model. Our approach to generate prompts for SAM from the relevance maps clearly outperforms this with a Dice of 0.33 after 6 iterations. This is already quite close to what we call the Dice with oracle of 0.35, where the threshold is determined per sample using the ground truth segmentation as a guide. Notable the MIL models showed less pronounced gains from postprocessing with SAM. 

\paragraph{Qualitative Results}

\begin{figure}[b]
    \centering
    \includegraphics[width=\textwidth]{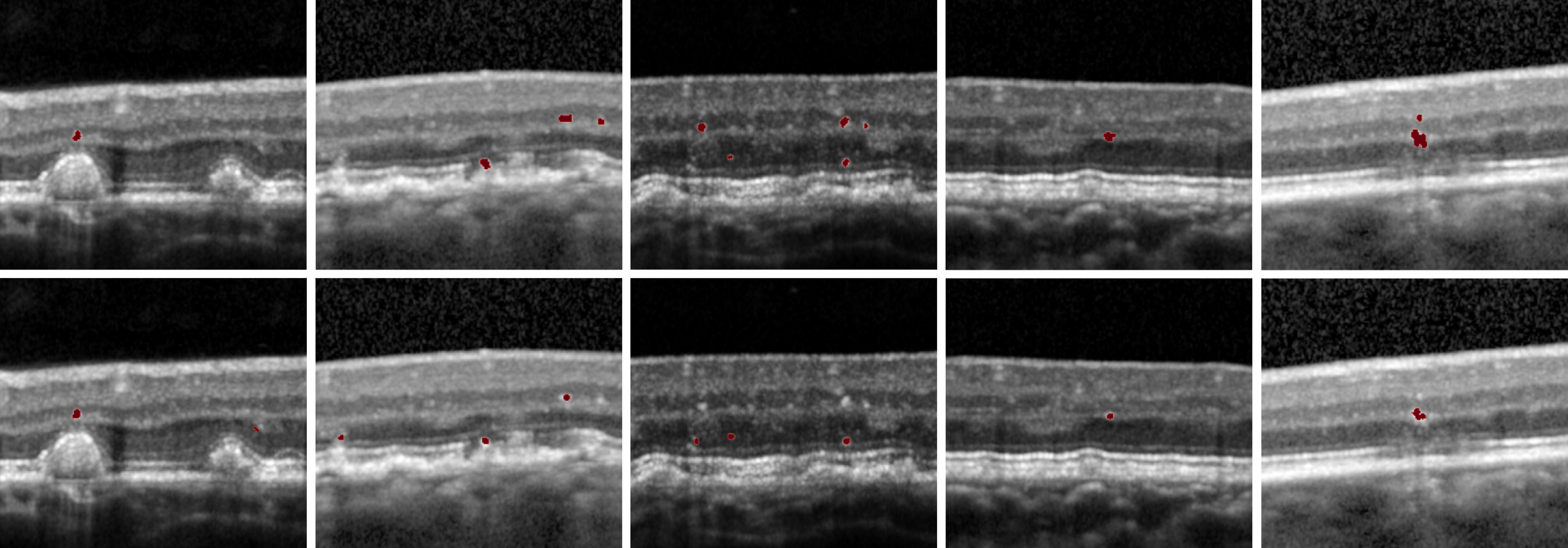}
    \caption{Example segmentations from the test set. The ground truth segmentation is shown on top and the output of our best performing weakly supervised segmentation model on the bottom.}
    \label{3696-fig:example_hrfs}
\end{figure}

In Fig. \ref{3696-fig:example_hrfs} we show example segmentations from the test set. One can see that our model successfully localizes several HRFs despite their small size.

\begin{table}[t]
\centering
\caption{Average Dice score for different postprocessings of the relevance maps. Threshold (Val) is a global threshold for the relevance maps calibrated on the validation set. Applying this threshold to the relevance maps from the validation and test set result in the Val Dice and Test Dice columns. The Oracle Test Dice is the average of each samples best possible Dice based on searching through a range of plausible thresholds.}
\label{3696-tab:SAM_Ablation}
\begin{tabular*}{\textwidth}{l@{\extracolsep\fill}rccccc}
\hline
\textbf{Model}       & \textbf{Task} & \textbf{Threshold (Val)} & \textbf{Val Dice} & \textbf{Test Dice} & \textbf{Oracle Test Dice} & \textbf{SAM Dice}  \\ 
\hline
\multirow{3}{*}{\textbf{CCT}} & 2-Classes     & 0.0037                   & 0.16              & 0.16               & 0.35                                & 0.33               \\
                     & 3-Classes     & 0.0065                   & 0.14              & 0.11               & 0.31                                & 0.31               \\
                     & Regression    & 0.0001                   & 0.06              & 0.06               & 0.21                                & 0.21               \\
\multirow{3}{*}{\textbf{MIL}} & 2-classes     & 0.0025                   & 0.11              & 0.11               & 0.19                                & 0.13               \\
                     & 3-Classes     & 0.0013                   & 0.10              & 0.03               & 0.15                                & 0.08               \\
                     & Regression    & 0.0007                   & 0.05              & 0.01               & 0.08                                & 0.07               \\
\hline
\end{tabular*}
\end{table}

\section{Discussion}
Regarding annotation style, it was our original hypothesis that training a model to predict image-level labels that merely indicate the presence of HRF might only be sufficient to detect a single (most prominent) HRF, while training with more information-rich labels that provide a more nuanced measure of HRF load might enable a more complete detection of all HRFs.

Our experiments, in which additional HRFs are segmented by masking out the ones that have already been found, indicate that a simple binary annotation, labeling B-scans as HRF positive/negative, is sufficient, and no additional benefit is gained from more time-consuming weak annotation styles.

Our strategy for prompting SAM~2 based on the most relevant pixel proved effective for obtaining final segmentation masks, achieving much better results than a simple thresholding of the relevance maps. Comparing Dice scores to those from an oracle-based thresholding that makes use of the ground truth suggests that we cannot expect to outperform SAM~2 even with sophisticated threshold adaptation.

Overall, our study highlights the effectiveness of CCT in conjunction with LRP and SAM~2 for the weakly supervised segmentation of HRFs in OCT, and we expect that it might be similarly effective for other segmentation problems that involve small structures. In the future, we plan to further refine our iterative inference, with the goal of achieving an even more favorable balance of recall and precision.

\begin{acknowledgement}
We would like to express our sincere gratitude to our collaborators, Zhichao Wu and Robyn Guymer, whose expertise and time were invaluable in assembling the dataset. We also thank our annotators, Gabriela Guzmann, Simon Janzen, and Kai Lyn Goh, for their efforts, as well as Shekoufeh Gorgi Zadeh for her technical guidance.
\end{acknowledgement}

% This command generates the bibliography using the entries of the .bib file.
% Remove it only if you do not use a bibliography. 
\printbibliography

@article{bachPixelWiseExplanationsNonLinear2015,
	title = {On {{Pixel-Wise Explanations}} for {{Non-Linear Classifier Decisions}} by {{Layer-Wise Relevance Propagation}}},
	author = {Bach, Sebastian and Binder, Alexander and Montavon, Gr{\'e}goire and Klauschen, Frederick and M{\"u}ller, Klaus-Robert and Samek, Wojciech},
	year = {2015},
	month = jul,
	journal = {PLoS One},
	volume = {10},
	number = {7},
	pages = {e0130140},
	publisher = {Public Library of Science},
	issn = {1932-6203},
	doi = {10.1371/journal.pone.0130140},
	urldate = {2024-11-04},
	langid = {english}
}

@article{hassaniEscapingBigData2022,
	author       = {Ali Hassani and
	Steven Walton and
	Nikhil Shah and
	Abulikemu Abuduweili and
	Jiachen Li and
	Humphrey Shi},
	title        = {Escaping the Big Data Paradigm with Compact Transformers},
	journal      = {CoRR},
	volume       = {abs/2104.05704},
	year         = {2021},
	url          = {https://arxiv.org/abs/2104.05704},
	eprinttype    = {arXiv},
	eprint       = {2104.05704},
	bibsource    = {dblp computer science bibliography, https://dblp.org}
}

@inproceedings{ilseAttentionbasedDeepMultiple2018,
	title = {Attention-Based {{Deep Multiple Instance Learning}}},
	booktitle = {Proceedings of the 35th {{International Conference}} on {{Machine Learning}}},
	author = {Ilse, Maximilian and Tomczak, Jakub and Welling, Max},
	year = {2018},
	month = jul,
	pages = {2127--2136},
	publisher = {PMLR},
	issn = {2640-3498},
	urldate = {2024-11-04},
	langid = {english}
}

@article{liPrevalenceIncidenceAgerelated2020,
	title = {Prevalence and Incidence of Age-Related Macular Degeneration in {{Europe}}: A Systematic Review and Meta-Analysis},
	shorttitle = {Prevalence and Incidence of Age-Related Macular Degeneration in {{Europe}}},
	author = {Li, Jeany Q and Welchowski, Thomas and Schmid, Matthias and Mauschitz, Matthias Marten and Holz, Frank G and Finger, Robert P},
	year = {2020},
	month = aug,
	journal = {British Journal of Ophthalmology},
	journal = {Br J Ophthalmol},
	volume = {104},
	number = {8},
	pages = {1077--1084},
	issn = {0007-1161, 1468-2079},
	doi = {10.1136/bjophthalmol-2019-314422},
	urldate = {2023-05-12},
	langid = {english}
}

@article{liuTSSKNetWeaklySupervised2023,
	title = {{{TSSK-Net}}: {{Weakly}} Supervised Biomarker Localization and Segmentation with Image-Level Annotation in Retinal {{OCT}} Images},
	author = {Liu, Xiaoming and Liu, Qi and Zhang, Ying and Wang, Meng and Tang, Jinshan and Liu, Xiaoming and Liu, Qi and Zhang, Ying and Wang, Meng and Tang, Jinshan},
	year = {2023},
	doi = {10.1016/J.COMPBIOMED.2022.106467},
	journal = {Computers in Biology and Medicine},
	journal = {Comput Biol Med},
	pages = {106467}
}

@article{maMSCAMMultiScaleClass2020a,
	title = {{{MS-CAM}}: {{Multi-Scale Class Activation Maps}} for {{Weakly-Supervised Segmentation}} of {{Geographic Atrophy Lesions}} in {{SD-OCT Images}}},
	shorttitle = {{{MS-CAM}}},
	author = {Ma, Xiao and Ji, Zexuan and Niu, Sijie and Leng, Theodore and Rubin, Daniel L. and Chen, Qiang},
	year = {2020},
	month = dec,
	journal = {IEEE Journal of Biomedical and Health Informatics},
	journal = {IEEE J Biomed Health Inform},
	volume = {24},
	number = {12},
	pages = {3443--3455},
	issn = {2168-2194, 2168-2208},
	doi = {10.1109/JBHI.2020.2999588},
	urldate = {2024-11-04},
	copyright = {https://ieeexplore.ieee.org/Xplorehelp/downloads/license-information/IEEE.html}
}

@article{raviSAM2Segment,
	author       = {Nikhila Ravi and
	Valentin Gabeur and
	Yuan{-}Ting Hu and
	Ronghang Hu and
	Chaitanya Ryali and
	Tengyu Ma and
	Haitham Khedr and
	Roman R{\"{a}}dle and
	Chlo{\'{e}} Rolland and
	Laura Gustafson and
	Eric Mintun and
	Junting Pan and
	Kalyan Vasudev Alwala and
	Nicolas Carion and
	Chao{-}Yuan Wu and
	Ross B. Girshick and
	Piotr Doll{\'{a}}r and
	Christoph Feichtenhofer},
	title        = {{SAM} 2: Segment Anything in Images and Videos},
	journal      = {CoRR},
	volume       = {abs/2408.00714},
	year         = {2024},
	url          = {https://doi.org/10.48550/arXiv.2408.00714},
	doi          = {10.48550/ARXIV.2408.00714},
	eprinttype    = {arXiv},
	eprint       = {2408.00714},
	bibsource    = {dblp computer science bibliography, https://dblp.org}
}

@article{vermaRelationshipDistributionIntraretinal2023,
	title = {Relationship between the Distribution of Intra-Retinal Hyper-Reflective Foci and the Progression of Intermediate Age-Related Macular Degeneration},
	author = {Verma, Aditya and Corradetti, Giulia and He, Ye and Nittala, Muneeswar G. and Nassisi, Marco and Velaga, Swetha B. and Haines, Jonathan L. and {Pericak-Vance}, Margaret A. and Stambolian, Dwight and Sadda, SriniVas R.},
	year = {2023},
	month = dec,
	journal = {Graefe's Archive for Clinical and Experimental Ophthalmology},
	journal = {Graefes Arch Clin Exp Ophthalmol},
	volume = {261},
	number = {12},
	pages = {3437--3447},
	issn = {1435-702X},
	doi = {10.1007/s00417-023-06180-4},
	urldate = {2024-11-04},
	langid = {english}
}

@article{wangWeaklySupervisedAnomaly2021,
	title = {Weakly Supervised Anomaly Segmentation in Retinal {{OCT}} Images Using an Adversarial Learning Approach},
	author = {Wang, Jing and Li, Wanyue and Chen, Yiwei and Fang, Wangyi and Kong, Wen and He, Yi and Shi, Guohua},
	year = {2021},
	month = aug,
	journal = {Biomedical Optics Express},
	journal = {Biomed Opt Express},
	volume = {12},
	number = {8},
	pages = {4713},
	issn = {2156-7085, 2156-7085},
	doi = {10.1364/BOE.426803},
	urldate = {2024-11-04},
	langid = {english}
}

@article{zhouFoundationModelGeneralizable2023,
	title = {A Foundation Model for Generalizable Disease Detection from Retinal Images},
	author = {Zhou, Yukun and Chia, Mark A. and Wagner, Siegfried K. and Ayhan, Murat S. and Williamson, Dominic J. and Struyven, Robbert R. and Liu, Timing and Xu, Moucheng and Lozano, Mateo G. and {Woodward-Court}, Peter and Kihara, Yuka and Altmann, Andre and Lee, Aaron Y. and Topol, Eric J. and Denniston, Alastair K. and Alexander, Daniel C. and Keane, Pearse A.},
	year = {2023},
	month = oct,
	journal = {Nature},
	volume = {622},
	number = {7981},
	pages = {156--163},
	publisher = {Nature Publishing Group},
	issn = {1476-4687},
	doi = {10.1038/s41586-023-06555-x},
	urldate = {2024-02-07},
	copyright = {2023 The Author(s)},
	langid = {english}
}

@InProceedings{Ali-LRP-transformer,
	title = 	 {{XAI} for Transformers: Better Explanations through Conservative Propagation},
	author =       {Ali, Ameen and Schnake, Thomas and Eberle, Oliver and Montavon, Gr{\'e}goire and M{\"u}ller, Klaus-Robert and Wolf, Lior},
	booktitle = 	 {Proc.\ 39th International Conference on Machine Learning},
	pages = 	 {435--451},
	year = 	 {2022},
	editor_old = 	 {Chaudhuri, Kamalika and Jegelka, Stefanie and Song, Le and Szepesvari, Csaba and Niu, Gang and Sabato, Sivan},
	volume = 	 {162},
	month = 	 jul,
	publisher =    {PMLR}
}

@article{gohHYPERREFLECTIVEFOCINOT2024,
	title = {Hyperreflective Foci not Seen as Hyperpigmentary Abnormalities on Color Fundus Photographs in Age-Related Macular Degeneration},
	author = {Goh, Kai Lyn and Wintergerst, Maximilian W. M. and Abbott, Carla J. and Hadoux, Xavier and Jannaud, Maxime and Kumar, Himeesh and Hodgson, Lauren A. B. and Guzman, Gabriela and Janzen, Simon and family=Wijngaarden, given=Peter, prefix=van, useprefix=true and Finger, Robert P. and Guymer, Robyn H. and Wu, Zhichao},
	date = {2024-02-01},
	journaltitle = {Retina},
	volume = {44},
	number = {2},
	eprint = {37831941},
	eprinttype = {pmid},
	pages = {214--221},
	issn = {1539-2864},
	doi = {10.1097/IAE.0000000000003958},
	langid = {english}
}

@article{guymerSubthresholdNanosecondLaser2019,
	title = {Subthreshold {{Nanosecond Laser Intervention}} in {{Age-Related Macular Degeneration}}: {{The LEAD Randomized Controlled Clinical Trial}}},
	shorttitle = {Subthreshold {{Nanosecond Laser Intervention}} in {{Age-Related Macular Degeneration}}},
	author = {Guymer, Robyn H. and Wu, Zhichao and Hodgson, Lauren A. B. and Caruso, Emily and Brassington, Kate H. and Tindill, Nicole and Aung, Khin Zaw and McGuinness, Myra B. and Fletcher, Erica L. and Chen, Fred K. and Chakravarthy, Usha and Arnold, Jennifer J. and Heriot, Wilson J. and Durkin, Shane R. and Lek, Jia Jia and Harper, Colin A. and Wickremasinghe, Sanjeewa S. and Sandhu, Sukhpal S. and Baglin, Elizabeth K. and Sharangan, Pyrawy and Braat, Sabine and Luu, Chi D. and {Laser Intervention in Early Stages of Age-Related Macular Degeneration Study Group}},
	date = {2019-06},
	journaltitle = {Ophthalmology},
	shortjournal = {Ophthalmology},
	volume = {126},
	number = {6},
	eprint = {30244144},
	eprinttype = {pmid},
	pages = {829--838},
	issn = {1549-4713},
	doi = {10.1016/j.ophtha.2018.09.015},
	langid = {english}
}

\end{document}